# Incremental Tradeoff Resolution in Qualitative Probabilistic Networks

**Chao-Lin Liu and Michael P. Wellman**
University of Michigan AI Laboratory
Ann Arbor, Michigan 48109, USA
{chaolin, wellman}@umich.edu

## Abstract

Qualitative probabilistic reasoning in a Bayesian network often reveals tradeoffs: relationships that are ambiguous due to competing qualitative influences. We present two techniques that combine qualitative and numeric probabilistic reasoning to resolve such tradeoffs, inferring the qualitative relationship between nodes in a Bayesian network. The first approach incrementally marginalizes nodes that contribute to the ambiguous qualitative relationships. The second approach evaluates approximate Bayesian networks for bounds of probability distributions, and uses these bounds to determinate qualitative relationships in question. This approach is also incremental in that the algorithm refines the state spaces of random variables for tighter bounds until the qualitative relationships are resolved. Both approaches provide systematic methods for tradeoff resolution at potentially lower computational cost than application of purely numeric methods.

## 1 Introduction

Researchers in uncertain reasoning regularly observe that to reach a desired conclusion (e.g., a decision), full precision in probabilistic relationships is rarely required, and that in many cases purely qualitative information (for some conception of "qualitative") is sufficient (Goldszmidt 1994). In consequence, the literature has admitted numerous schemes attempting to capture various forms of qualitative relationships (Wellman 1994), useful for various uncertain reasoning tasks. Unfortunately, we generally lack a robust mapping from tasks to the levels of precision required, and indeed, necessary precision is inevitably variable across problem instances. As long as some potential problem might require precision not captured in the qualitative scheme, the scheme is potentially inadequate for the associated task. Advocates of qualitative uncertain reasoning typically acknowledge this, and sometimes suggest that one can always revert to full numeric precision when necessary. But specifying a numerically precise probabilistic model as a fallback preempts any potential model-specification benefit of the qualitative scheme, and so it seems that one may as well use the precise model for everything.[1] This is perhaps the primary reason that qualitative methods have not seen much use in practical applications of uncertain reasoning to date.

The case for qualitative reasoning in contexts where numerically precise models are available must appeal to benefits other than specification, such as computation. Cases where qualitative properties justify computational shortcuts are of course commonplace (e.g., independence), though we do not usually consider this to be qualitative reasoning unless some inference is required to establish the qualitative property itself in order to exploit it. Since pure qualitative inference can often be substantially more efficient than its numeric counterpart (e.g., in methods based on infinitesimal probabilities (Goldszmidt & Pearl 1992) or ordinal relationships (Druzdzel & Henrion 1993)), it is worth exploring any opportunities to exploit qualitative methods even where some numeric information is required.

We have begun to investigate this possibility for the task of deriving the qualitative relationship (i.e., the sign of the probabilistic association, defined below) between a pair of variables in a Bayesian network. From an abstracted version of the network, where all local relationships are described qualitatively, we can derive the entailed sign between the variables of interest efficiently using propagation techniques. However, since the abstraction process discards information, the result may be qualitatively ambiguous even if the actual relationship entailed by the precise

---

[1] If the qualitative formalism is a strict abstraction, then any conclusions produced by the precise model will agree at the qualitative level. Even in such cases, qualitative models may have benefits for explanation or justification (Henrion & Druzdzel 1991), as they can indicate something about the robustness of the conclusions (put another way, they can concisely convey broad classes of conclusions).

model is not.

In this paper, we report on two approaches that use qualitative reasoning to derive these relationships without necessarily resorting to solution of the complete problem at full precision, even in cases where purely qualitative reasoning would be ambiguous. Both approaches are incremental, in that they apply numeric reasoning to either subproblems or simplified versions of the original, to produce an intermediate model more likely to be qualitatively unambiguous.

The next section reviews the concepts of qualitative influences and tradeoffs in a network model. The third section explains the incremental marginalization approach, followed by the experimental results. We then discuss the state-space abstraction approach, and conclude with a brief comparison of our approaches with some others.

## 2 Qualitative probabilistic networks

### 2.1 Qualitative influences

Qualitative probabilistic networks (QPNs) (Wellman 1990) are abstractions of Bayesian networks, with conditional probability tables summarized by the signs of qualitative relationships between variables. Each arc in the network is marked with a sign—positive ($+$), negative ($-$), or ambiguous ($?$)—denoting the sign of the qualitative probabilistic relationship between its terminal nodes.

The interpretation of such qualitative influences is based on *first-order stochastic dominance (FSD)* (Fishburn & Vickson 1978). Let $F(x)$ and $F'(x)$ denote two cumulative distribution functions (CDFs) of a random variable $X$. Then $F(x)\ FSD\ F'(x)$ holds if and only if (iff)

$$F(x) \leq F'(x) \text{ for all } x. \qquad (1)$$

We say that one node positively influences another iff the latter's conditional distribution is increasing in the former, all else equal, in the sense of $FSD$.

**Definition 1 ((Wellman 1990))** *Let $F(z|x_i, y)$ be the cumulative distribution function of $Z$ given $X = x_i$ and the rest of $Z$'s parent nodes $Y = y$. We say that node $X$ positively influences node $Z$, denoted $S^+(X, Z)$, iff*

$$\forall x_i, x_j, y.\quad x_i \leq x_j \Rightarrow F(z|x_j, y)\ FSD\ F(z|x_i, y).$$

Analogously, we say that node $X$ *negatively influences* node $Z$, denoted $S^-(X, Z)$, when we reverse the direction of the dominance relationship in Definition 1. The arc from $X$ to $Z$ in that case carries a negative sign. When the dominance relationship holds for both directions, we denote the situation by $S^0(X, Z)$. However, this entails conditional independence, and so we typically do not have a direct arc from $X$ to $Z$ in this case. When none of the preceding relationships between the two CDFs hold, we put a question mark on the arc, and denote such situations as $S^?(X, Z)$. We may apply the preceding definitions to boolean nodes under the convention that **true** $>$ **false**.

### 2.2 Inference and tradeoff resolution

Given a QPN, we may infer the effects of the change in the value of one variable on the values of other variables of interest. The inference can be carried out via graph reduction (Wellman 1990), or qualitative propagation techniques (Druzdzel & Henrion 1993).

If we are fortunate, we may acquire decisive answers from the qualitative inference algorithms. Often, however, the results of such qualitative reasoning are ambiguous. This might be because the relationship in question actually is ambiguous (i.e., nonmonotone or context-dependent), or due to loss of information in the abstraction process.

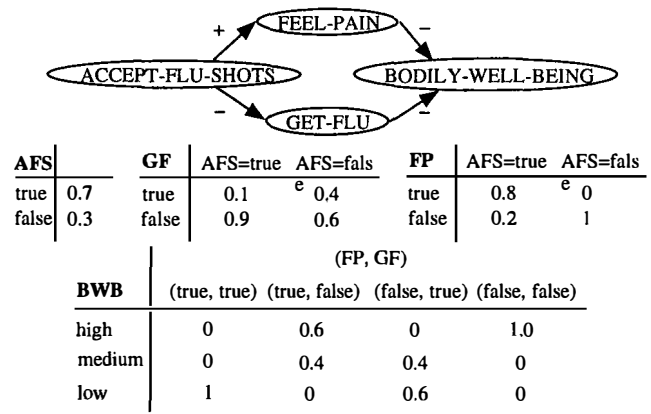

Figure 1: A simple case of qualitative ambiguity.

This can happen, for instance, when there are competing influential paths from the source node—whose value is tentatively modified—to the target node—whose change in value is of interest. While accept flu shots may decrease the probability of get flu, it also increases the probability and degree of feel pain. On the other hand, increasing either get flu or feel pain decreases overall bodily well-being, all else equal. As a result, qualitative reasoning about the problem of whether we should accept flu shots will yield only an ambiguous answer. The situation is illustrated by the QPN in Figure 1, where there is one positive path and one negative path from accept flu shots to bodily well-being. The combination of these two paths is qualitatively ambiguous. Worse, the ambiguity of this relationship would propagate within any network for which this pattern forms a subnetwork. For example, if this issue plays a role in a decision whether to go to a doctor, the result would be ambiguous regardless of the other variables involved.

Had we applied more precise probabilistic knowledge, such as a numerically specified Bayesian network, the result may have been decisive. Indeed, if accept flu shots and bodily well-being are binary, then a fully precise model is by necessity qualitatively unambiguous. However, performing all inference at the most precise level might squander some advantages of the qualitative approach. In the developments below, we consider some ways to apply numeric inference incrementally, to the point where qualitative reasoning can produce a decisive result.

We use the term *tradeoff resolution* to refer to the task of resolving the qualitatively ambiguous relationship between variables of interest. Next, we demonstrate that in the worst case, qualitative tradeoff resolution may be no easier than full numerical inference.

### 2.3 Computational complexity of tradeoff resolution

**Theorem 1** *Qualitative tradeoff resolution is NP-hard.*

We show that resolving qualitative tradeoff is an NP-hard problem by reducing the problem of computing absolute approximations to the task of qualitative tradeoff resolution. An estimate $\gamma$ is an absolute approximation of $\Pr(y)$ if

$$\Pr(y) - \delta \leq \gamma \leq \Pr(y) + \delta,$$

where $\delta$ is the range of error. The problem of computing absolute approximations has been shown NP-hard (Dagum & Luby 1993).

Consider the task of computing absolute approximations for $\Pr(y)$ in a given Bayesian network in which $Y$ is a boolean variable. We construct a corresponding tradeoff resolution problem for this task as follows. The network for this tradeoff resolution problem includes the given Bayesian network and two boolean variables $D$ and $T$. The tradeoff resolution task is to determine the qualitative influence of $D$ on $T$ in the network shown in Figure 2. The cloud where $Y$ resides represents the given Bayesian network. We use $x$ and $\bar{x}$ to denote that $X$ is true and false, respectively.

To check the overall influence of $D$ on $T$, we need to know whether $\Pr(t|d) \geq \Pr(t|\bar{d})$. Using the data shown in the following figure, we can show that the previous inequality implies that $\Pr(y) \leq \frac{1}{1+\varepsilon}$. Notice that the range of $\frac{1}{1+\varepsilon}$ is $[1/2, 1]$ when we change $\varepsilon$ between 1 and 0.

Therefore, using an efficient algorithm for tradeoff resolution, we can efficiently determine the range of $\Pr(y)$ if its range is in $[1/2, 1]$. This can be done by setting $\varepsilon$ to a very small number $\alpha$, and gradually setting $\varepsilon$ to multiples of $\alpha$, i.e., $2\alpha, 3\alpha, \ldots$, etc. As a result, when the range of $\Pr(y)$ is in $[1/2, 1]$, we will find a more precise range of $\Pr(y)$ by calling the efficient algorithm for tradeoff resolution $O(\frac{1}{\alpha})$ times. Similarly, we can determine the range of $\Pr(y)$ when the range is in $[0, 1/2]$, by calling the efficient algorithm for tradeoff resolution $O(\frac{1}{\alpha})$ times. This can be done by setting $\Pr(t|d,\bar{y})$ and $\Pr(t|\bar{d},y)$ to $\varepsilon$ and 1 in Figure 2, respectively.

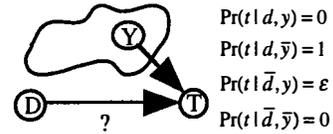

Figure 2: Resolving qualitative influence of $D$ on $T$ yields an approximate probability for $Y$.

## 3 Incremental marginalization

### 3.1 Node reduction

The idea of incremental marginalization is to reduce the network node-by-node until the result is qualitatively unambiguous. The basic step is Shachter's arc reversal operation.

**Theorem 2 ((Shachter 1988))** *If there is an arc from node $X$ to node $Y$ in the given Bayesian network, and no other directed paths from $X$ to $Y$, then we may transform the network to one with an arc from $Y$ to $X$ instead. In the new network, $X$ and $Y$ inherit each other's parent nodes.*

Let $\boldsymbol{P}_X$, $\boldsymbol{P}_Y$, and $\boldsymbol{P}_{XY}$ respectively denote $X$'s own parent nodes, $Y$'s own parent nodes, and $X$ and $Y$'s common parent nodes in the original network, and let $\boldsymbol{P}_{Y'} = \boldsymbol{P}_Y - \{X\}$. The new conditional probability distribution of $Y$ and $X$ are determined by the following:

$$\Pr^{new}(y|\boldsymbol{p}_X, \boldsymbol{p}_{Y'}, \boldsymbol{p}_{XY})$$
$$= \sum_X \Pr^{old}(y|\boldsymbol{p}_Y, \boldsymbol{p}_{XY}) \Pr^{old}(x|\boldsymbol{p}_X, \boldsymbol{p}_{XY})$$
$$\Pr^{new}(x|y, \boldsymbol{p}_X, \boldsymbol{p}_{Y'}, \boldsymbol{p}_{XY})$$
$$= \frac{\Pr^{old}(y|\boldsymbol{p}_Y, \boldsymbol{p}_{XY}) \Pr^{old}(x|\boldsymbol{p}_X, \boldsymbol{p}_{XY})}{\Pr^{new}(y|\boldsymbol{p}_X, \boldsymbol{p}_{Y'}, \boldsymbol{p}_{XY})}$$

On reversing all the outgoing arcs from node $X$, the node becomes *barren* and can be removed from the network. The net effect of reversing arcs and removing barren nodes as described is equivalent to marginalizing node $X$ from the network.

### 3.2 Tradeoff resolution via node reduction

Consider the QPN shown on the left-hand side of Figure 3. Since there exist both a positive path (through $X$) and a negative path (direct arc) from $W$ to $Z$, the qualitative influence of $W$ on $Z$ is ambiguous. This local "?" would propagate throughout the network, necessarily ambiguating the relationship of any predecessor of $W$ to any successor of $Z$.

Once we have detected the source of such a local ambiguity, we may attempt to resolve it by marginalizing node $X$.

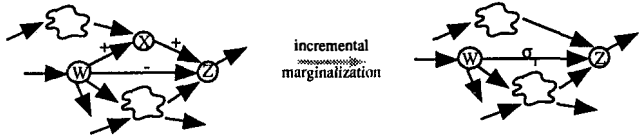

Figure 3: Marginalizing $X$ potentially resolves the qualitative influence of $W$ on $Z$.

The new sign on the direct arc from $W$ to $Z$ can be determined by inspecting the new conditional probability table of $Z$, given by Equation (2). If we are fortunate, the qualitative sign $\sigma_1$ may turn out to be decisive, in which case we have resolved the tradeoff.

This example illustrates the main idea of the incremental marginalization approach to resolving tradeoffs in QPNs. If we obtain an unambiguous answer to the desired qualitative relationship from the reduced network after marginalizing a selected node, then there is no need to do further computation. If the answer is still ambiguous, we may select other nodes to marginalize. The iteration continues until a decisive answer is uncovered. We present the skeleton of the Incremental TradeOff Resolution algorithm below. The algorithm is designed to answer queries about the qualitative influence of a decision node on a target node, using a given strategy for selecting the next node to reduce.

**Algorithm 1** *ITOR(decision, target, strategy)*

1. *Remove nodes that are irrelevant to the query about decision's influence on target (Shachter 1988).*

2. *Attempt to answer the query via qualitative inference (Druzdzel & Henrion 1993).*

3. *If the answer to the query is decisive, exit; otherwise continue.*

4. *Select a node to reduce according to strategy. If there is no node that can be reduced, return "?", else perform the node reduction, and calculate the qualitative abstractions of the transformed relationships. Return to Step 2.*

We expect the incremental approach to improve performance over purely numeric inference on average. Since qualitative inference is quadratic whereas exact inference in Bayesian networks is exponential in the worst case, the qualitative inference steps do not add appreciably to computation time. On the other hand, when the intermediate results suffice to resolve the tradeoff, we save numeric computation over whatever part of the network is remaining.

### 3.3 Prioritizing node reduction operations

The objective of carrying out node-reduction operations in ITOR is to resolve qualitative tradeoffs. The optimal strategies for respective tasks will differ, in general. For example, a node that is very expensive to reduce at a certain stage of the evaluation might be the best prospect for resolving the tradeoff.

We exploit intermediate information provided in qualitative belief propagation (Druzdzel & Henrion 1993) in determining which node to reduce next. If we can propagate a decisive qualitative influence from the decision node $D$ all the way to the target node $T$, we will be able to answer the query. Otherwise, there must be a node $X$ that has an indecisive relationships from $D$. Recall that we have pruned nodes irrelevant to the query, so any nodes that have indecisive relationship with $D$ will eventually make the relationship between $D$ and $T$ indecisive. We have identified several conceivable strategies based on this observation, and have tried two of them thus far.

The first strategy is to reduce node $X$, as long as $X$ is not the target node $T$. When $X$ is actually $T$, we choose to reduce the node $Y$ that passed the message to $X$ changing its qualitative sign from a decisive one to "?". However, this $Y$ cannot be $D$ itself. If it is, then either (1) there are only two nodes remaining in the network, and there is no decisive answer to the query, or (2) there are other nodes, and we randomly pick among those adjacent to $D$ or $T$.

The second strategy is similar to the first, except that we exchange the priority of reducing $X$ and $Y$. We handle the situations where $X$ and/or $Y$ happen to be $D$ and/or $T$ in the same manner as in the first strategy.

These strategies have the advantage that finding the next node to reduce does not impose extra overhead in the ITOR algorithm. The selection is a by-product of the qualitative inference algorithm. However neither of these strategies (nor any that we know) is guaranteed to minimize the cost of resolving the tradeoff.

## 4 Experimental study

We have tested the effectiveness of the algorithm using randomly generated network instances. The experiments are designed to examine how connectivity of the network, sizes of state spaces, and strategies for scheduling node reduction affect the performance of the algorithm.

### 4.1 Generating random networks

In the experiments, we use Bayesian networks in which arcs can be assigned decisive qualitative signs. To this end, we construct QPNs with only decisive signs on arcs, and then use the signs to govern the way we assign conditional probability values for nodes in their corresponding Bayesian network. The conditional probability distributions of nodes and the qualitative signs on arcs must agree with each other.

To create a random QPN with $n$ nodes and $l$ arcs, we first create a complete directed acyclic graph (DAG) with $n$ nodes. Each arc in this DAG is assigned a random number that is sampled from a uniform distribution. We then attempt to remove the arc with the largest assigned number, under the constraint that the DAG remains connected. If removing the arc with the largest assigned number will make the DAG disconnected, we will attempt to remove the arc with the next largest number. We remove arcs until the DAG contains only $l$ arcs. After creating the network structure, we randomly assign qualitative signs (positive or negative) to the arcs.

We then build a Bayesian network that corresponds to the generated QPN, that is, respects its structure and qualitative signs. We select the cardinality of each node by sampling from a uniform distribution over the range $[2, MC]$, where $MC$ denotes the maximum state-space cardinality. For nodes without parents, we assign prior probabilities by selecting parameters from a uniform distribution and then normalizing.

For a node $X$ with parent nodes $PA(X)$, the qualitative signs in the QPN dictate a partial ordering of the conditional probability distributions for various values of $X$, where the distributions are ordered based on the $FSD$ relationship. Let $pa_i(X)$ denote an instantiation of the parent nodes of $X$. To enforce this ordering, we identify the $pa_i(X)$ that requires us to make the distribution $F(x|pa_i(X))$ dominate distributions $F(x|pa_j(X))$ for all other $pa_j(X)$. We assign the parameters for $F(x|pa_i(X))$ (as for priors) by sampling from a uniform distribution. We then assign the remaining distributions in stages, at each stage setting only those distributions dominated by the previously assigned distributions. We make these assignments using the same random procedure, but under the constraint that the resulting distribution must respect the qualitative signs given the previous assignments.

### 4.2 Results

In each experiment, we specify the number of nodes, the number of arcs, and maximum cardinality of state spaces for the randomly generated networks. In all experiments, we create networks with 10 nodes before pruning. We query the qualitative influence from the node1 to node10, and disregard the instances in which the answer is ambiguous after exact evaluation of the network.

Since the first step of the ITOR algorithm prunes nodes irrelevant to the query, the network actually used in inference is usually simpler than the original network. In Table 1, we record the *average* number of nodes and links after the pruning step. $MC$ denotes maximum cardinality. All experiments reported used the first node selection strategy; results from the second strategy were virtually identical.

| nodes | links | MC | $R_{nodes}$ | $R_{reversals}$ |
|---|---|---|---|---|
| 8.0 | 14.2 | 2 | 0.697 | 0.722 |
| 8.0 | 14.4 | 3 | 0.730 | 0.754 |
| 9.2 | 26.1 | 2 | 0.846 | 0.869 |
| 9.4 | 26.8 | 3 | 0.855 | 0.874 |

Table 1: Experimental results. Each experiment runs ITOR over 10000 random networks with decisive influence from node1 to node10.

We measure the performance of ITOR with two metrics. The first metric, $R_{nodes}$, is the ratio of the number of reduced nodes when the decisive answer is found to the number of nodes that would be reduced in exact numerical evaluation. The second metric, $R_{reversals}$, is the ratio of number of arc reversal operations already done when the solution is found to the number of arc reversal operations that would be carried out for exact numerical evaluation. The latter figure is based on an arbitrary strategy for reducing the remaining network after the tradeoff is resolved, however, and so would tend to be an optimistic estimate of the saving. Table 1 reports averages for each metric. The savings due to incremental tradeoff resolution are $1 - R_{nodes}$ and $1 - R_{reversals}$, respectively, and so lower values of the metrics indicate better performance.

The results in Table 1 suggest that ITOR offers greater performance for sparsely connected networks and smaller state spaces. Further experimentation may lead us to more precise characterization of the expected savings achievable through incremental marginalization.

## 5 State-space abstraction

Recall that qualitative relationships are defined based on the concept of first-order stochastic dominance defined by (1). Since only qualitative relationships among variables are of interest, exact calculation of the values of the CDFs may not be necessary if we can use approximate CDFs to determine whether $FSD$ holds.

In previous work, we report an iterative state-space abstraction (ISSA) algorithm for approximate evaluation of Bayesian networks (Wellman & Liu 1994). The ISSA algorithm aggregates states of selected variables, called *abstracted nodes*, into *superstates* to construct abstract versions of the *original Bayesian networks* (OBNs) that specify exact probability distributions. The aggregation of states requires the reassignment of the conditional probability tables (CPTs) of both the abstracted nodes and their child nodes. We call the method used in this assignment task a CPT assignment *policy*. These abstract Bayesian networks (ABNs) are then used to compute point-valued approximations of the probability distributions of interest. As a result of the abstraction operations, the computational cost of evaluating the ABNs can be less than that of evaluating the OBNs. The algorithm iteratively refines the state spaces of the selected abstracted nodes for improving the quality of approximations, when the allocated computation time permits.

In this section, we introduce a new policy for computing bounds of probability distributions, and apply the revised ISSA algorithm to resolving ambiguous qualitative relationships.

### 5.1 Motivation and definitions

Consider the task of determining whether $F(x|d_i)$ dominates $F(x|d_j)$. Assume that we have ways to control approximation methods to obtain approximate CDFs $\hat{F}(x|d_i)$ and $\hat{F}(x|d_j)$ such that $F(x|d_i) \leq \hat{F}(x|d_i)$ and $\hat{F}(x|d_j) \leq F(x|d_j)$ for all $x$. Given these approximate CDFs, $F(x|d_i)$ $FSD$ $F(x|d_j)$ will hold if we also have $\hat{F}(x|d_i) \leq \hat{F}(x|d_j)$ for all $x$. In other words, it is possible to determine qualitative relationship using bounds of probability distributions. We define bounds of CDFs as follows.

**Definition 2** *A CDF $\overline{F}(x)$ is an upper bound of $F(x)$, if $F(x) \leq \overline{F}(x)$ for all $x$. A CDF $\underline{F}(x)$ is a lower bound of $F(x)$, if $\underline{F}(x) \leq F(x)$ for all $x$.*

In terms of these definitions, $S^-(D, X)$ holds if there exist $\overline{F}(x|d_i)$ and $\underline{F}(x|d_j)$ such that

$$\text{for all } x, \ d_i < d_j \Rightarrow \overline{F}(x|d_i) \leq \underline{F}(x|d_j). \quad (2)$$

Similarly, $S^+(D, X)$ holds if there exist $\overline{F}(x|d_j)$ and $\underline{F}(x|d_i)$ such that

$$\text{for all } x, \ d_i < d_j \Rightarrow \overline{F}(x|d_j) \leq \underline{F}(x|d_i). \quad (3)$$

In addition, we may be able to tell that $D$ neither positively nor negatively influences $T$ by examining bounds. Specifically, a sufficient condition for $S^?(D, X)$ is that there exist $x_r$, $x_s$, and bounds such that, for some $d_i < d_j$,

$$\overline{F}(x_r|d_i) < \underline{F}(x_r|d_j) \text{ and } \overline{F}(x_s|d_j) < \underline{F}(x_s|d_i). \quad (4)$$

When (4) holds, the curves for $F(x|d_i)$ and $F(x|d_j)$ must intersect as illustrated in the following figure.

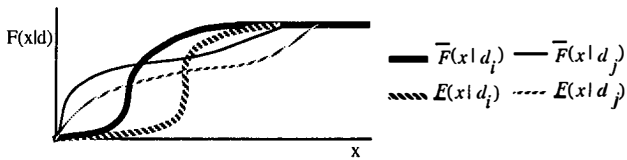

Figure 4: Bounds of $F(x|d)$ imply $S^?(D, X)$.

### 5.2 Bounding probability distributions

We may compute bounds of conditional probability distributions by using an ISSA algorithm that applies the *dominance policy* in aggregating states of abstracted nodes (Liu & Wellman 1998). Let $\boldsymbol{PA}(A)$ be the set of parent nodes of an abstracted node $A$, and $[a_{i,j}]$ the superstate representing the aggregation of states from $a_i$ through $a_j$, $i \leq j$. The dominance policy modifies the CPT of $A$ as follows:

$$\hat{\Pr}([a_{i,j}]|\boldsymbol{pa}(A)) = \sum_{k=i}^{j} \Pr(a_k|\boldsymbol{pa}(A)).$$

Let $Y$ be a child node of $A$, and $\boldsymbol{PX}(Y)$ be the subset of parent nodes of $Y$ excluding $A$. Depending on whether we want to compute lower or upper bounds of selected CDFs, we *strengthen* or *weaken* the conditional probability distribution of $Y$ given its parent nodes. To strengthen the distribution, we assign $\hat{F}(y|[a_{i,j}], \boldsymbol{px}(Y))$ as follows:

$$\hat{F}(y|[a_{i,j}], \boldsymbol{px}(Y)) = \min_{l \in [i,j]} F(y|a_l, \boldsymbol{px}(Y)).$$

To weaken the distribution, we assign $\hat{F}(y|[a_{i,j}], \boldsymbol{px}(Y))$ as follows:

$$\hat{F}(y|[a_{i,j}], \boldsymbol{px}(Y)) = \max_{l \in [i,j]} F(y|a_l, \boldsymbol{px}(Y)).$$

We have identified and reported conditions under which the ISSA algorithm may compute bounds of conditional probability distributions (Liu & Wellman 1998). Taking advantage of qualitative relationship and conditional independence among variables, we can compute lower and upper bounds of desired conditional probability distributions.

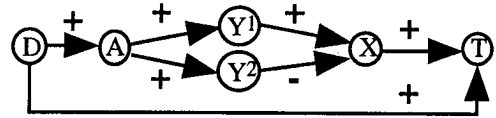

Figure 5: We may use qualitative relationships for bounding probability distributions.

Consider the network in Figure 5. We have (a) $Y^1$ positively influences $X$ given $D$ and $Y^2$, (b) $Y^2$ negatively influences $X$ given $D$ and $Y^1$, (c) $X$ and $A$ are independent given $D$, $Y^1$, and $Y^2$, (d) $[D, A, Y^1, Y^2]$ is an ancestral ordering, and (e) $Y^1$ is not a descendant of $Y^2$ and vice versa in the network. An ordering $[J^1, J^2, \ldots, J^n]$ of nodes in a set of nodes $\boldsymbol{J}$ is an ancestral ordering if, for every $J^i \in \boldsymbol{J}$, all the ancestors of $J^i$ are ordered before $J^i$. Given these conditions, we can compute the bounds of $F(x|d)$ when we abstract $A$ with the dominance policy. Specifically, we obtain lower (upper) bounds of $F(x|d)$ by weakening (strengthening) $F(y^1|a, y^2)$ and strengthening (weakening) $F(y^2|a, y^1)$ with respect to $A$ when we abstract $A$. In addition, We can obtain lower (upper) bounds of $F(x|d)$ by strengthening (weakening) $F(x|y^1)$ with respect to $Y^1$ when we abstract $Y^1$, given that (a) $[D, Y^1, X]$ is an ancestral ordering and (b) $X$ is the only child node of $Y^1$. Analogously, we may abstract $Y^2$ in computing bounds of $F(x|d)$ to further reduce computation time.

In addition, we have shown that bounds computed by the ISSA algorithm tighten as we refine the state space of the abstracted nodes (Liu & Wellman 1998). Therefore, we are more likely to resolve qualitative tradeoffs as we carry out more iterations of the ISSA algorithm. The computation can terminate whenever we determine the qualitative relationship of interest.

### 5.3 Tradeoff resolution via approximation

As we discussed in Section 3.3, if the relationship between $D$ and $T$ is ambiguous, there must be a node $X$ such that $X$ is marked with "?" when we propagate the sign from $D$ toward $T$. As an alternative to incremental marginalization, we may apply any approximate evaluation algorithm for Bayesian networks in Step 4 of ITOR, if the approximation algorithm can return bounds of conditional probability distributions. We use ISSA with the dominance policy as such an alternative in the following discussion.

Using the ISSA algorithm with dominance policy can save computation time for the tradeoff resolution task. As mentioned, the ISSA algorithm may find the correct qualitative relationship by the time it needs to exactly evaluate $F(x|d)$ using the conditions specified in (2) to (4). In addition, the ISSA algorithm may compute the bounds of $F(x|d)$ by evaluating a portion of the given Bayesian network. For instance, $T$ in Figure 5 is barren and can be ignored for the computation of $F(x|d)$.

Notice that we should not terminate ITOR when ISSA returns a "?" for the relationship of $D$ and $X$ in Step 4. When this occurs, we need to continue ITOR as usual. It is possible that we find a decisive relationship between $D$ and $T$ even when some nodes in the network have ambiguous relationship with $D$. For instance, using ISSA, we might find that $D$ positively influences $T$ even if $D$ neither negatively nor positively influences $X$ in the network shown in the following figure. Therefore, ITOR should run until either a decisive relationship between $D$ and $T$ has been found in Step 2 or an ambiguous relationship between $D$ and $T$ is confirmed by ISSA in Step 4.

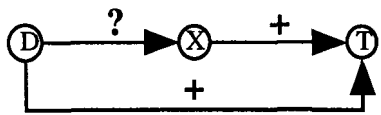

Figure 6: ITOR should not terminate when ISSA reports $S^?(D, X)$.

Some results obtained by ISSA may be reused in later iterations of ITOR. Take the network in Figure 5 as an example. Assume that we find that $D$ negatively influences $X$ in Figure 5 by exactly computing the values of $F(x|d)$, and that we are about to use the ISSA algorithm to determine the qualitative relationship between $D$ and $T$. In this case, if we do not reuse previous results, we may abstract $A$, $Y^1$, $Y^2$, and $X$ in computing the bounds of $F(t|d)$ in Step 4. However, given that $CI(T, \{D, X\}, \{A, Y^1, Y^2\})$ and that we have computed the exact values of $F(x|d)$, the network has been reduced to the one shown in the following figure. Therefore, we could save computation time by running ISSA over the network in Figure 7, and should not run ISSA over the network in Figure 5 from scratch.

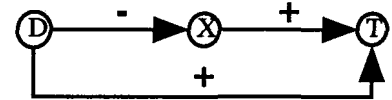

Figure 7: Reduced version of the network in Figure 5.

Whether we reuse the information about the bounds that are obtained from running ISSA is a design issue. For instance, assume that we find that $D$ negatively influences $X$ in Figure 5 by applying (2). Given this result, the qualitative relationship between $D$ and $T$ is still ambiguous since there are still two competing influential paths from $D$ to $T$ as indicated in Figure 7. Upon locating this ambiguity, ITOR uses ISSA to resolve the ambiguity, and the issue is whether we reuse the information we have about $F(x|d)$ that is obtained from the previous execution of ISSA. Given that $CI(T, \{D, X\}, \{A, Y^1, Y^2\})$ and that $D$ decisively influences $X$ in the network, a heuristic is that we use the knowledge about $F(x|d)$ obtained in the previous run of ISSA first. To do so, we set the conditional probability of $X$ given $D$ according to bounds of $F(x|d)$ in the network shown in Figure 7, and use this approximate network for resolving the ambiguity. If this opportunistic approach does not resolve the qualitative ambiguity between $D$ and $T$, we then use ISSA to evaluate a network that includes $A$, $Y1$ and $Y2$. Another alternative is to directly compute bounds of $F(t|d)$ using the original network. It is possible that we can resolve the ambiguity when $A$, $Y1$, $Y2$, and $X$ have very small number of states in ISSA. The optimal choice for this design issue varies from network to network, depending on their underlying probability distributions.

## 6 Discussion

We have shown that resolving qualitative tradeoff is NP-hard, and have discussed the application of incremental marginalization and state-space abstraction methods to the qualitative tradeoff resolution task. The incremental marginalization approach iteratively reduces a node in the network, and the state-space abstraction approach approximately evaluates a portion of the network. Initial exper-iments with incremental marginalization suggest that noticeable savings are possible, but definitive evaluation of both methods awaits further empirical and theoretical investigation.

The incremental marginalization approach bears some similarity to symbolic probabilistic inference, as in the *variable elimination* (VE) algorithm (Zhang & Poole 1996), in that we sum out one node from the Bayesian network at a time. The ITOR algorithm differs from the VE algorithm in the determination of the elimination ordering, and of course in the stopping criterion.

Parsons and Dohnal (1993) discuss a semiqualitative approach for inference using Bayesian networks. The basic idea is similar to state-space abstraction. The center of their work is to design calculus for computing the the probability intervals of variables, and their methods may work even when the conditional probabilities in Bayesian networks are not completely specified. However, their methods cannot be applied to the qualitative tradeoff resolution task. Parsons (1995) also exploits more special numerical relationships among variables to define new qualitative relationships for refining inference in QPN.

There are other approaches that make use of numerically specified knowledge in qualitative inference. For instance, Kuipers and Berleant (1988) apply incompletely specified numerical information in qualitative inference tasks.

The incremental approaches we propose in this paper provide systematic ways to resolving qualitative tradeoffs at potentially lower computational cost than fully precise methods. Empirical results suggest that incremental marginalization can provide savings for some networks. How to use qualitative information to guide the scheduling of node reduction and how to reuse partial results obtained from approximate evaluation of Bayesian networks to achieve the best performance possible remain as open problems for future work.